\documentclass{article}

\usepackage{PRIMEarxiv}

\usepackage[utf8]{inputenc} % allow utf-8 input
\usepackage[T1]{fontenc}    % use 8-bit T1 fonts
\usepackage{hyperref}       % hyperlinks
\usepackage{url}            % simple URL typesetting
\usepackage{booktabs}       % professional-quality tables
\usepackage{amsfonts}       % blackboard math symbols
\usepackage{nicefrac}       % compact symbols for 1/2, etc.
\usepackage{microtype}      % microtypography
\usepackage{lipsum}
\usepackage{wrapfig}
\usepackage{cleveref}
\usepackage{enumitem}
\usepackage{graphicx}
\usepackage{makecell}
\usepackage{pifont}
\usepackage{fancyhdr}       % header
\usepackage{graphicx}       % graphics
\graphicspath{{media/}}     % organize your images and other figures under media/ folder
\newcommand{\cmark}{\ding{51}}
\newcommand{\xmark}{\ding{55}}
\usepackage[table]{xcolor} %投稿前确定是否支持
\definecolor{lightpurple}{RGB}{242,230,242}
\usepackage{orcidlink}
\usepackage{hyperref}

\usepackage{multirow} %用于合并单元格
\newcommand{\eg}{e.g.\ }
\newcommand{\ie}{i.e.\ }

\newcommand{\vs}{vs.\ }
\newcommand{\OurModelName}{\textsc{Cheers}} %模型名字

\title{\texorpdfstring{\OurModelName{}\includegraphics[width=0.035\textwidth]{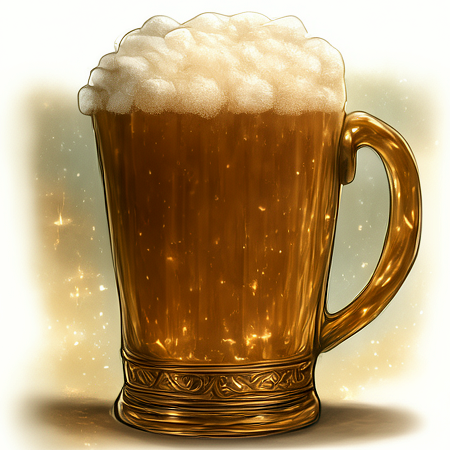}}{\OurModelName}: Decoupling Patch Details from Semantic Representations Enables Unified Multimodal Comprehension and Generation}

\author{\textbf{Yichen Zhang$^{1}$}\thanks{Equal contribution.}\quad\textbf{Da Peng$^{2}$}\footnotemark[1]\quad\textbf{Zonghao Guo$^{1}$}\thanks{Corresponding author.}\quad\textbf{Zijian Zhang$^3$}\quad\textbf{Xuesong Yang$^3$}\\\textbf{Tong Sun$^3$}\quad\textbf{Shichu Sun$^3$}\quad\textbf{Yidan Zhang$^3$}\quad\textbf{Yanghao Li$^{1}$}\quad\textbf{Haiyan Zhao$^{1}$}\quad\textbf{Wang Xu$^{1}$}\\\textbf{Qi Shi$^{1}$}\quad\textbf{Yangang Sun$^{1}$}\quad\textbf{Chi Chen$^{1}$}\quad\textbf{Shuo Wang$^{1}$}\quad\textbf{Yukun Yan$^{1}$}\quad\textbf{Xu Han$^{1}$}\\\textbf{Qiang Ma$^{1}$}\quad\textbf{Wei Ke$^{2}$}\quad\textbf{Liang Wang$^3$}\quad\textbf{Zhiyuan Liu$^{1}$}\quad\textbf{Maosong Sun$^{1}$}\\
$^1$Tsinghua University\quad
$^2$Xi'an Jiaotong University\quad 
$^3$University of Chinese Academy of Sciences \\
{\textit{yichen0zhang@gmail.com}\quad\textit{guozonghao96@outlook.com}\quad\textit{metapda@gmail.com}}\\
\begin{tabular}{llll}
\huggingface&\url{\hflink}&\github&\url{\ghlink}\\
\end{tabular}
}

\begin{document}
\maketitle
\begin{abstract}
  % 研究现状
  A recent cutting-edge topic in multimodal modeling is to unify visual comprehension and generation within a single model. 
  % 面临问题：任务所需视觉表示的mismatching 导致难以联合优化
  However, the two tasks demand mismatched decoding regimes and visual representations, making it non-trivial to jointly optimize within a shared feature space.
  % 直接引入我们的模型
  In this work, we present \OurModelName{}, a unified multimodal model that decouples patch-level details from semantic representations, thereby stabilizing semantics for multimodal understanding and improving fidelity for image generation via gated detail residuals.
  % 介绍模块
  \OurModelName{} includes three key components: (i) a unified vision tokenizer that encodes and compresses image latent states into semantic tokens for efficient LLM conditioning, (ii) an LLM-based Transformer that unifies autoregressive decoding for text generation and diffusion decoding for image generation, and (iii) a cascaded flow matching head that decodes visual semantics first and then injects semantically gated detail residuals from the vision tokenizer to refine high-frequency content.
  % 取得的效果
  Experiments on popular benchmarks demonstrate that \OurModelName{} {\color{black}matches or surpasses} advanced UMMs in both visual understanding and generation. 
  % \OurModelName{} also achieves 4$\times$ token compression, enabling more efficient high-resolution image encoding and generation. 
  Notably, \OurModelName{} outperforms the Tar-1.5B on the popular benchmarks GenEval and MMBench, while requiring only 20\% of the training cost, indicating effective and efficient (\ie, 4$\times$ token compression) unified multimodal modeling.
  We will release all code and data for future research. 
  \keywords{Unified multimodal model \and Visual generation and comprehension \and Unified vision encoder.}
\end{abstract}
\section{Introduction}
\label{sec:intro}
\begin{figure}[t]
    \centering
    \includegraphics[width=\linewidth]{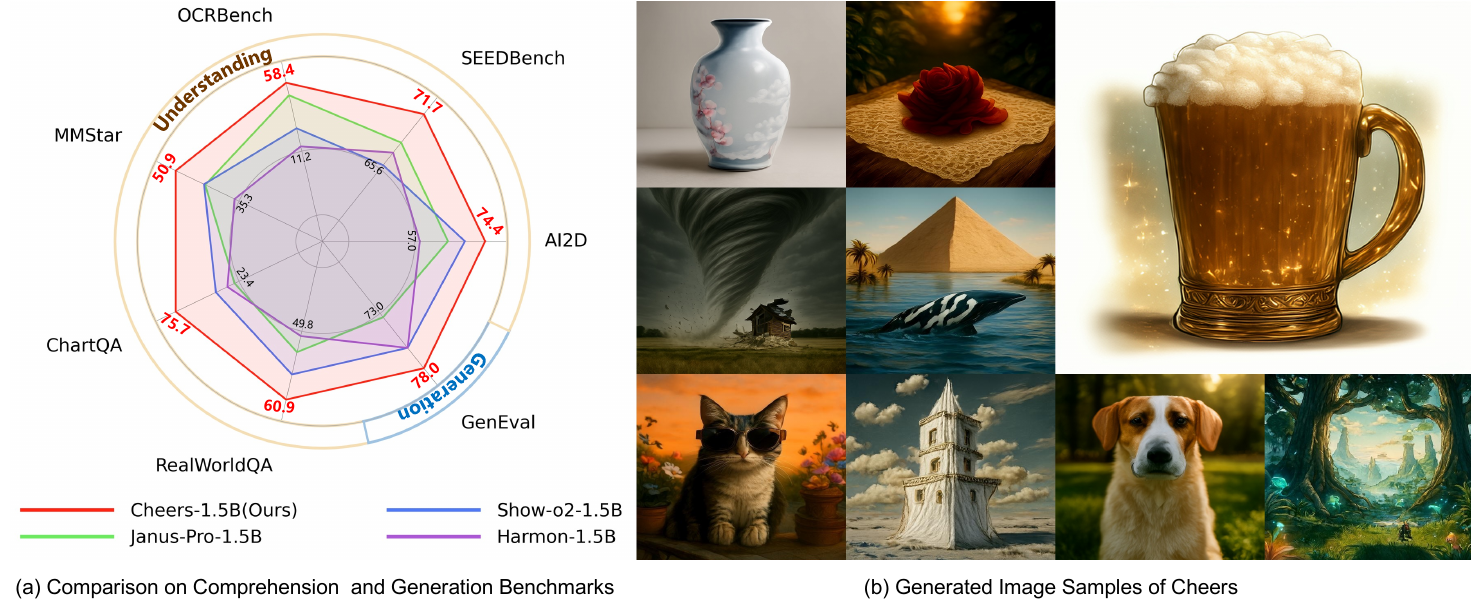}
    \caption{\OurModelName{} Capabilities. (a)~Performance on general understanding and generation benchmarks compared with unified multimodal models (UMMs) of similar scale. (b)~Generated image samples of \OurModelName{}.}
    \label{fig:head}
\end{figure}

% 入题写清楚mllms和dits的现状

\noindent Multimodal large language models (MLLMs)~\cite{bai2025qwen2,wang2025internvl3,zhang2025videollama,team2026kimi} have largely matured for visual comprehension, while diffusion models~\cite{zheng2025diffusion, flux-2-2025, Peebles2022DiT, rombach2021highresolution, lin2024sdxl, labs2025flux1kontextflowmatching} have set the standard for high-fidelity image generation. Bringing both into a single model is a cutting-edge step toward more human-like multimodal intelligence.
% 说问题
However, such unification is particularly challenging, as the two tasks demand fundamentally different decoding mechanisms and visual representations.

In terms of decoding mechanisms, discretizing visual representations~\cite{
van2017neural,han2025tar,tian2024visual,han2024clip} for autoregressive (AR) prediction with text tokens offers a seamless adaptation to existing MLLM architectures~\cite{song2025dualtoken,liu2024world,zhou2024transfusion}. 
However, discrete tokens suffer from quantization errors~\cite{Chameleon_Team_Chameleon_Mixed-Modal_Early-Fusion_2024, xie2024sana} and dimensional constraints~\cite{zhangdimensional, song2025dualtoken,Chameleon_Team_Chameleon_Mixed-Modal_Early-Fusion_2024}, leading to the loss of visual information. Bypassing the constraints of sequential raster-scanning of image generation, recent approaches~\cite{xie2025show,liu2025tuna,deng2025emerging} integrate diffusion modeling to capture global visual context alongside AR-based text generation.
%

\begin{figure}[t!]
\centering
\includegraphics[width=\linewidth]{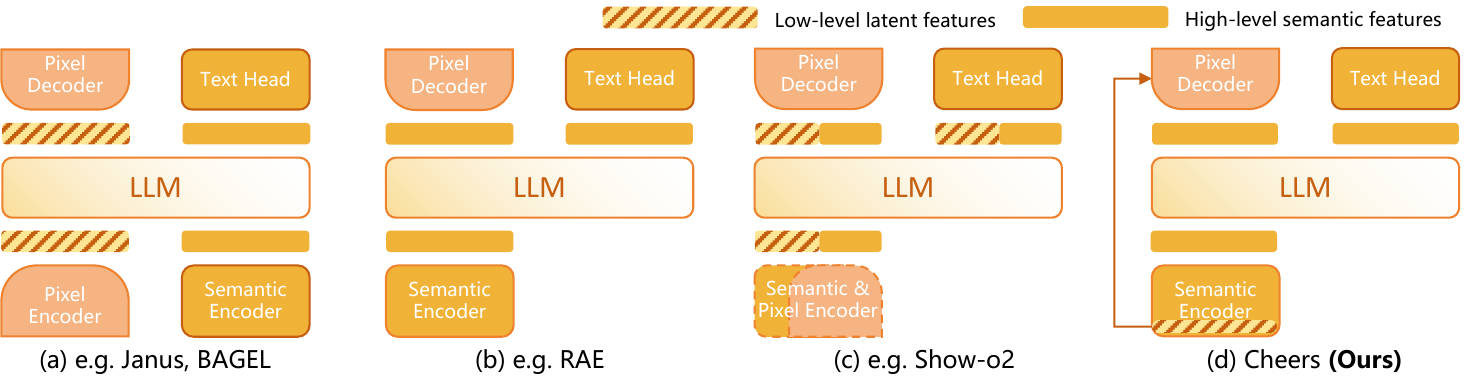}
\caption{Architectural comparison between prior UMMs and \OurModelName{}. (a)~Separated visual spaces for understanding and generation. (b)~Single semantic-centric space with limited structural details. (c)~Fused feature representation with potential interference. (d)~Cheers (Ours): A unified vision tokenizer that integrates structural and semantic features to ensure stable semantic understanding while enhancing generative details.}
\label{fig:architectural_comparison}
\end{figure}

% 逐一回答两个问题
From the perspective of visual representations, multimodal understanding typically relies on semantic-rich features from vision encoders~\cite{radford2021learning,tschannen2025siglip}, whereas high-fidelity image generation often depends on detail-preserving latents from reconstruction-oriented tokenizers~\cite{kingma2013auto,van2017neural,yao2025reconstruction}. 
However, relying solely on a single representation often fails to simultaneously satisfy these distinct requirements~\cite{sun2023emu,sun2024generative,wang2024emu3,cui2025emu3,xie2024show,Chameleon_Team_Chameleon_Mixed-Modal_Early-Fusion_2024,team2025nextstep}, as shown in~\cref{fig:architectural_comparison}~(b).
% 列举现状
Therefore, one line of UMMs~\cite{deng2025emerging,wu2024janus,chen2025janus} separates the feature optimization for visual comprehension and generation, achieving strong task-specific performance, as shown in~\cref{fig:architectural_comparison}~(a). 
The other line seeks to integrate these capabilities via a unified token interface by either fusing heterogeneous features~\cite{xie2025show,liao2025mogao} or jointly optimizing a shared vision tokenizer with multiple objectives~\cite{qu2025tokenflow,wu2024vila,yao2025towards,du2025vqrae}, as shown in~\cref{fig:architectural_comparison}~(c). 

Despite these inspiring explorations, the intrinsic optimization conflict between visual comprehension and generation remains insufficiently investigated in UMMs. 
In this paper, we introduce \OurModelName{}, a UMM that decouples patch-level details from semantic representations, stabilizing semantics for image understanding and improving generation fidelity by injecting high-frequency detail residuals, as shown in~\cref{fig:architectural_comparison}~(d).
% 介绍核心模块
\OurModelName{} includes three key components. (i) A unified vision tokenizer utilizes a representation encoder (\eg, SigLIP2-ViT) upon VAE latents to extract semantic features, subsequently compressed via a pixel-unshuffle~\cite{wang2025internvl3} operation for efficient LLM conditioning.
(ii) An LLM-based Transformer integrates autoregressive and diffusion decoding for text and image generation, respectively, thereby capitalizing on the superior modeling paradigms inherent to each modality. 
(iii) At the core of \OurModelName{} is a cascaded flow matching head that explicitly decouples image generation into two phases: it initially synthesizes high-level semantics at a low resolution, followed by injecting semantically gated high-frequency residuals from the vision tokenizer to achieve precise super-resolution generation. This is akin to painting, where global structure precedes fine-grained detailing, a perspective that resonates with recent work~\cite{baade2026latent}.

Extensive experiments on standard benchmarks demonstrate that \OurModelName{} performs on par with or exceeds state-of-the-art UMMs in both visual comprehension and generation, validating the efficacy of our unified modeling approach. 
\OurModelName{} also represents a significant step towards token-compressed UMMs, achieving a 4$\times$ compression rate for efficient high-resolution image understanding and generation.
Notably, \OurModelName{} outperforms the Tar model on the popular benchmarks GenEval and MMBench, while requiring only 20\% of the training cost, indicating effective and efficient unified multimodal modeling.

% {\color{red}Intriguingly, we observe that unifying visual representations within a semantic space fosters the emergence of image editing capabilities, even in the absence of explicit supervision for such tasks.}

Our contribution can be summarized as threefold. (1) We propose decoupling patch details from semantic representations, which redefine the multimodal feature modeling trajectory of UMMs, alleviating the optimization interference between comprehension and generation tasks. (2) We introduce \OurModelName{}, a hybrid-decoding UMM equipped with a unified vision tokenizer that achieves significant token compression for efficient multimodal modeling. (3) We perform extensive evaluations on popular benchmarks to verify the effectiveness of \OurModelName{}, providing detailed analysis and insights for future research.

%` \section{Our Method: \OurModelName{}\includegraphics[width=0.035\textwidth]{figures/CHEERS.png}}
% \section{\OurModelName{}\includegraphics[width=0.035\textwidth]{figures/CHEERS.png}}
\section{\texorpdfstring{\OurModelName{}\includegraphics[width=0.035\textwidth]{figures/CHEERS.png}}{\OurModelName}}
\label{sec:framework}

% In this section, we present the overall framework of our proposed method, \OurModelName{}. Specifically, \cref{sec:architecture} introduces the model architecture, \cref{sec:inference} presents the inference strategy along with the training objectives, and \cref{sec:training_pipeline} details the training procedure.
We present the \OurModelName{} framework, covering its architecture (\cref{sec:architecture}), inference strategy and objectives (\cref{sec:inference}), and training pipeline (\cref{sec:training_pipeline}).

\subsection{Model Architecture}
\label{sec:architecture}
\begin{figure}[tb]
\centering
\includegraphics[width=\linewidth]{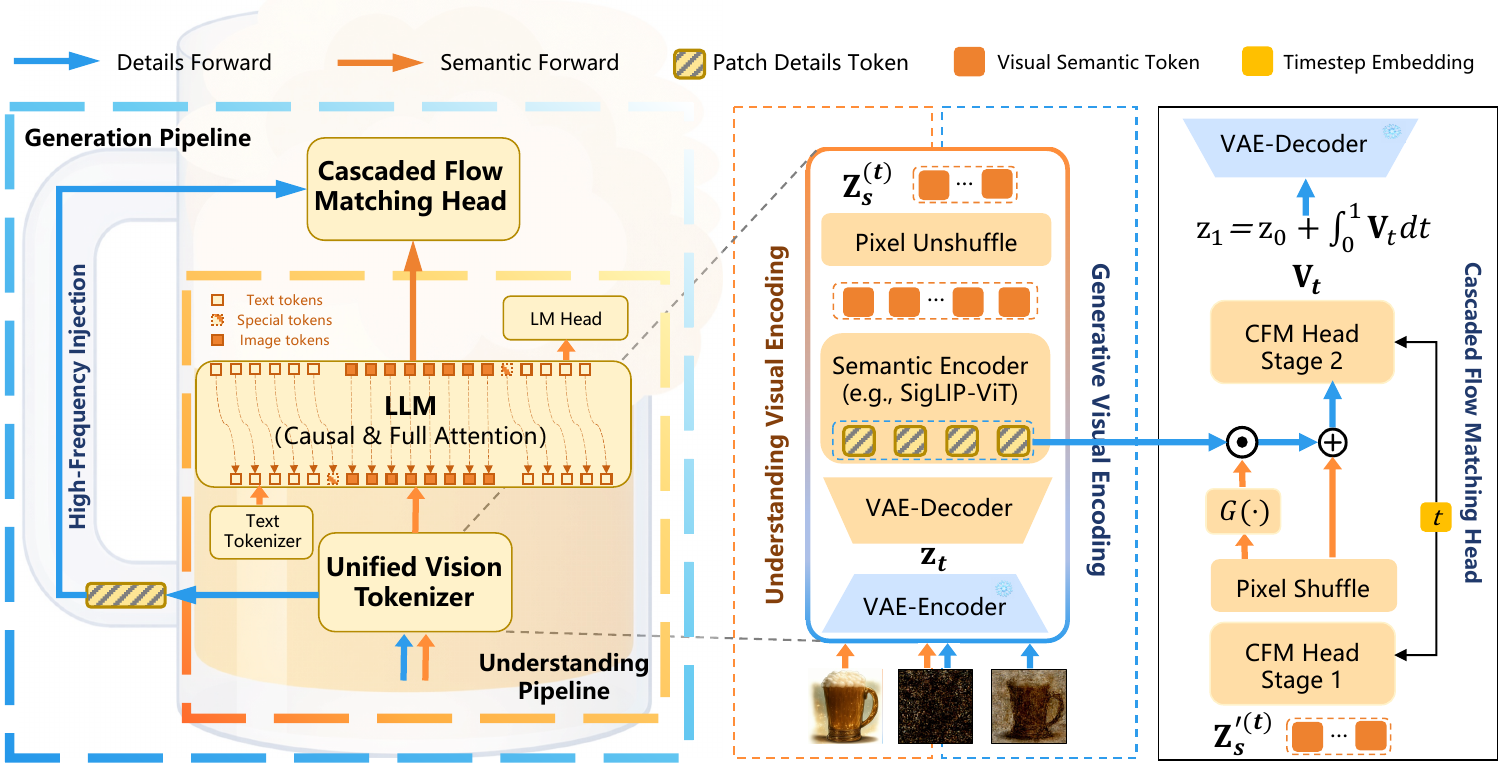}
\caption{Overview of \OurModelName{}, a unified framework for multimodal understanding and image generation. The Unified Vision Tokenizer converts visual inputs into semantic tokens that are jointly processed with text tokens by the LLM for understanding tasks, and detail tokens that serve as step-adaptive high-frequency injection into the CFM Head during generation. During generation, the CFM Head predicts a continuous-time velocity field in the latent space, enabling iterative sampling from Gaussian noise $\mathbf{z}_{0}$ to the terminal latent $\mathbf{z}_{1}$, which is finally decoded by the VAE decoder.}
\label{fig:model_framework}
\end{figure}

As illustrated in~\cref{fig:model_framework}, \OurModelName{} is built upon three key components: a unified vision tokenizer for visual encoding, a unified LLM-based Transformer backbone for multimodal modeling, and a cascaded flow matching head for image generation. Additionally, a standard text tokenizer and a language modeling (LM) head are employed for language encoding and text generation, respectively, to support visual understanding tasks.

\noindent\textbf{Unified Vision Tokenizer.} 
% As shown in~\cref{fig:model_framework}, \OurModelName{} leverages a semantic encoder, \ie, SigLIP2-ViT~\cite{tschannen2025siglip} to extract unified visual representations based on the latent states produced by the VAE encoder~\cite{flux-2-2025}.
As illustrated in~\cref{fig:model_framework}, \OurModelName{} adopts a unified vision tokenizer composed of a VAE decoder~\cite{flux-2-2025} and a semantic encoder, \ie, SigLIP2-ViT~\cite{tschannen2025siglip}. Specifically, the latent representations produced by the VAE encoder are first decoded into the image space via the VAE decoder, after which SigLIP2 extracts high-level semantic visual features. In this way, the VAE decoder and SigLIP2 jointly function as an integrated module that bridges latent representations and unified semantic visual embeddings.

Specifically, given an input image $\mathbf{X} \in \mathbb{R}^{H \times W \times 3}$, where $H$ and $W$ denote the image height and width, we first process it through a VAE encoder, yielding the latent states $\mathbf{z}_1 \in \mathbb{R}^{h \times w \times d}$, where $h = H/16$, $w = W/16$, and $d$ represents the latent feature dimension. 
To unify diverse tasks, we formulate a task-dependent latent $\mathbf{z}_t = t\mathbf{z}_1 + (1 - t)\mathbf{z}_0$, where latent noise $\mathbf{z}_0 \sim \mathcal{N}(0, 1)$. We sample timestep $t \in (0, 1)$ for image generation, fix $t = 1$ (\ie, $\mathbf{z}_t = \mathbf{z}_1$) for visual understanding, and set $t = 0$ (\ie, $\mathbf{z}_t = \mathbf{z}_0$) for language-only tasks.
Subsequently, instead of directly processing these latent states $\mathbf{z}_t$ using a ViT with randomly initialized patch embeddings like~\cite{liu2025tuna,xie2025show}, $\mathbf{z}_t$ is passed through a VAE decoder $D(\cdot)$ to reconstruct the pixel-level image. The reconstructed image is then encoded by the ViT backbone to extract high-level semantic tokens $\mathbf{z}_{s}^{(t)} \in \mathbb{R}^{h \times w \times d'}$, where $d'$ denotes the semantic feature dimension. To ensure strict spatial alignment between these semantic tokens and the latent patches, we adopt SigLIP2-ViT with a 16$\times$16 patch embedding layer $S(\cdot)$. 
Notably, we experimentally found that direct latent processing like~\cite{liu2025tuna} discards fine-grained features and hinders OCR-centric understanding ability. By reconstructing the pixel space, we circumvent this issue and successfully retain essential visual details. {\color{black}Please kindly refer to the Supplement for details.}

Before feeding the semantic tokens into our unified LLM-based Transformer, a Pixel-Unshuffle module~\cite{wang2025internvl3} is applied to reduce their spatial resolution and project the channel dimension, resulting in $\mathbf{Z}_{s}^{(t)} \in \mathbb{R}^{h/2 \times w/2 \times c}$, where $c$ is the LLM hidden size. 
To the best of our knowledge, we are the first work to introduce 2D token compression within a UMM.

\noindent\textbf{Unified LLM-based Transformer.}
To achieve optimal image-text joint modeling, we utilize autoregressive decoding for text generation and diffusion processes for image generation within a single LLM backbone, \ie, Qwen2.5-1.5B-Instruct~\cite{qwen2.5}.
Specifically, given the semantic visual tokens $\mathbf{Z}_s^{(t)}$ and the text embeddings $\mathbf{Z}_{text}$ derived from input instructions via the text tokenizer, we concatenate them into a unified input sequence, which is then processed by the LLM backbone to yield contextualized hidden states through deep cross-modal encoding. 
Note that a bidirectional attention mask is applied to $\mathbf{Z}_s^{(t)}$ to capture global visual context, whereas a causal mask is employed for $\mathbf{Z}_{text}$ to enable AR decoding.
Depending on the task modality, the LLM outputs are subsequently routed to different decoding paradigms.
For visual comprehension or pure text generation, the model employs a standard AR language modeling objective. For image generation, the continuous visual hidden states $\mathbf{Z}_{s}^{(t)}$, which have been integrated with the text instructions or descriptions $\mathbf{Z}_{text}$, are decoded via our cascaded flow matching head. 

\noindent\textbf{Cascaded Flow Matching Head.}
Inspired by~\cite{fan2025prism, baade2026latent, chen2025unihetero}, we propose to explicitly decouple high-frequency visual details from low-frequency semantic features and then integrate them during image synthesis.
Specifically, our CFM head consists of two cascaded stages, comprising 7 and 3 DiT blocks~\cite{Peebles2022DiT} respectively. Both stages employ the AdaLN-Zero~\cite{Peebles2022DiT} architecture to incorporate temporal modulations of the denoising procedure from the timestep $t$.
In the first stage, the CFM head takes the contextualized hidden states $\mathbf{Z}_{s}^{(t)} \in \mathbb{R}^{h/2 \times w/2 \times c}$ from the LLM as input to perform low-resolution semantic generation. This is followed by a PixelShuffle~\cite{shi2016real} module that up-samples the feature maps to 2$\times$ resolution and low-dimension ones $\mathbf{Z'}_s^{(t)} \in \mathbb{R}^{h \times w \times d'}$.
In the second stage, given the high-frequency patch details $S(D(\mathbf{z}_t)) \in \mathbb{R}^{h \times w \times d'}$, we first introduce a gating network $G(\cdot)$ to adaptively control the injection of fine-grained information to update the decoded features $\mathbf{Z'}_s^{(t)}$ as 
\[\mathbf{Z'}_s^{(t)} \leftarrow G(\mathbf{Z'}_s^{(t)}) \odot S(D(\mathbf{z}_t)) + \mathbf{Z'}_s^{(t)},\]
where $G(\mathbf{Z'}_s^{(t)}) \in \mathbb{R}^{h \times w \times 1}$ denotes a scalar map and $\odot$ the element-wise multiplication. Notably, as $\mathbf{Z'}_s^{(t)}$ is modulated by the timestep $t$ in the first stage, the intensity of high-frequency injection (HFI) is dynamically coupled with the generative trajectory. Our empirical analysis (see~\cref{sec:visualization}) reveals that, even without explicit supervision, the magnitude of HFI naturally intensifies as $t$ progresses.

Finally, $\mathbf{Z'}_s^{(t)}$ is fed into subsequent DiT layers to predict the velocity field $\mathbf{V}_t$. 
Such a progression mirrors the hierarchical nature of human drawing, which naturally transitions from global layout sketching to localized detail refinement.

\subsection{Inference and Training Objectives}
\label{sec:inference}
\noindent\textbf{Inference.}
% For pure text understanding and multimodal understanding, we follow the standard practice of sequentially sampling tokens from the predicted distribution. For image generation, we use classifier-free guidance (CFG), similar to prior works~\cite{xie2025show}.
For text-only and multimodal understanding tasks, we follow standard autoregressive decoding by sequentially selecting tokens from the predicted distribution. For image generation, we perform continuous-time flow-based sampling starting from Gaussian noise in the latent space, denoted as $\mathbf{z}_{0}$. At each time step $t$, we feed the current latent variable $\mathbf{z}_{t}$ into the unified vision tokenizer to obtain the corresponding visual tokens, which are then jointly processed with the textual condition by the LLM. Subsequently, the CFM head predicts the continuous-time velocity field $\mathbf{V}_{t}$ based on the LLM outputs, and we update the latent state via numerical integration:
\[\mathbf{z}_{t+\Delta t}=\mathbf{z}_t+\int_{t}^{t+\Delta t} \mathbf{V}_{\tau}\, d\tau .\]
% The updated latent $Z_{t+\Delta t}$ is subsequently treated as the new state and fed back into the Unified Vision Tokenizer for the next iteration. We iterate the above procedure---Tokenizer $\rightarrow$ LLM $\rightarrow$ CLM Head $\rightarrow$ ODE update---until reaching the terminal latent $Z_{1}$. Finally, $Z_{1}$ is decoded by the VAE Decoder to produce the output image.
The updated latent variable $\mathbf{z}_{t+\Delta t}$ serves as the input to the next integration step. By repeatedly applying tokenization, conditional modeling with the LLM, velocity prediction through the CFM Head, and numerical ODE integration, the latent trajectory is evolved from $\mathbf{z}_{0}$ to the terminal state $\mathbf{z}_{1}$. The final latent $\mathbf{z}_{1}$ is then decoded using the VAE decoder to produce the output image.

In addition, following prior work~\cite{xie2025show}, we adopt classifier-free guidance (CFG) during generation. To further adjust the time noise schedule in flow-based sampling, we apply a schedule shift and rescale the continuous-time variable with a hyperparameter $\alpha$. Formally, given the original time step $t \in [0,1]$, the shifted time step is computed as
$\tilde{t}=\frac{\alpha t}{1+(\alpha-1)t}$.
% We set $\alpha=1$ by default and $\alpha=0.05$ For DPG-Bench~\cite{hu2024ella}.

\noindent\textbf{Training Objectives.}
We use an end-to-end unified training optimization. For visual comprehension or pure text generation, the probability of generating the target text sequence $\mathbf{y} = \{y_1, \dots, y_L\}$ is factorized as $P_\theta(\mathbf{y} | \mathbf{C}) = \prod_{i=1}^L p_\theta(y_i | \mathbf{y}_{<i}, \mathbf{C})$, where $\mathbf{C}$ represents the conditioning context, \ie, $[\mathbf{Z}_{s}^{(t)}]$ for image caption, $[\mathbf{Z}_{s}^{(t)}, \mathbf{Z}_{text}]$ for image question-answer or prefix $[\mathbf{Z}_{text}]$ for pure text. 
We use the standard cross-entropy loss function $\mathcal{L}_{AR} = -\log P_\theta(\mathbf{y} | \mathbf{C})$, where $\mathbf{y}$ is the generated target text sequence, $\mathbf{C}$ is the conditioning context, and $\theta$ are the learnable parameters of the model. For the image generation part, we use the flow matching loss function $\mathcal{L}_{FM} = \| v_{\theta}(\mathbf{Z'}_s^{(t)}) - (\mathbf{z}_1 - \mathbf{z}_0) \|_2^2$. The overall training loss is the weighted sum of the text loss and image generation loss, given by:
\[\mathcal{L}_{total} = \mathcal{L}_{AR} + \lambda \mathcal{L}_{FM}\]
where $\lambda$ is a hyperparameter used to balance the loss between text generation and image generation, and in our training, $\lambda$ is set to $1$.

\subsection{Training Pipeline}
\label{sec:training_pipeline}
\begin{table}[t]
\caption{Training setup and hyperparameters across different stages for \OurModelName{}.}
\label{tab:training_details}
\centering
\resizebox{\textwidth}{!}{
\setlength{\tabcolsep}{2pt} % 默认6pt，可根据需要调成3~4pt
\begin{tabular}{@{\hspace{2pt}}l|c|c|c|c@{\hspace{2pt}}}
\toprule
\textbf{Stage}&\textbf{Vision-Language Alignment}&\textbf{General Pre-Training}&\textbf{Refined Pre-Training}&\textbf{Supervised Fine-Tuning}\\
\midrule
\multirow{4}{*}{\makecell{Dataset}}
&\multirow{4}{*}{\makecell{Image caption\\Text-to-image}}
&\multirow{4}{*}{\makecell{Pure text corpus\\Detailed image caption\\Text-to-image\\OCR data}}
&\multirow{4}{*}{\makecell{Pure text corpus\\Synthetic generation data\\General VQA}}
&\multirow{4}{*}{\makecell{Pure text corpus\\Instruction text-to-image\\High-quality instruction data}}
\\
&&&&
\\
&&&&
\\
&&&&
\\
\cmidrule(lr){1-5}
Learning rate&$1e^{-4}$&$1e^{-4}$&$4e^{-5}$&$2e^{-5}$\\
Training parts&Projection \& CFM Head \& Gate&All parameters w/o VAE&All parameters w/o VAE &All parameters w/o VAE\\
Schedules&constant&constant&constant&cosine\\
% Gradient clip&1.0&1.0&1.0&1.0\\
% Optimizer&AdamW&AdamW&AdamW&AdamW\\
% Warmup ratio&0.02&0.02&0.02&0.02\\
Batch size&512&512&512&128\\
Training steps&{30 K}&{60 K}&{65 K}&{30 K}\\
\bottomrule
\end{tabular}
}
\end{table}
\noindent Our four-stage progressive training is detailed in~\cref{tab:training_details}. Image resolution is fixed at $512\times512$. We initialize the image encoder from Siglip2 and FLUX.2. All experiments use the AdamW optimizer with a 0.02 warmup ratio and 1.0 gradient clipping, conducted on 128 NVIDIA A100 GPUs (16 nodes).

\noindent\textbf{Stage I: Vision–Language Alignment.} We train only the randomly initialized modules (projector, CFM head, and gating modules). The training data consists of 4.5M image-caption pairs from the LLaVA-UHD-v3~\cite{sun2025llava} and 1.3M ImageNet samples re-annotated by Qwen2.5-VL-3B~\cite{bai2025qwen2}. To establish preliminary generative capability, we repeat the ImageNet dataset 10 times.

\noindent\textbf{Stage II: General Pre-Training.} Subsequently, we optimize all model parameters except the VAE using 30M multimodal samples. Understanding data comprises captions from Infinity-MM~\cite{gu2024infinity}, LLaVA-UHD-v3~\cite{sun2025llava}, and TextAtlas5M~\cite{wang2025textatlas5m}. Generation data, including pretraining data from BLIP-3o~\cite{chen2025blip3}, and a small portion of synthetic data re-generated using FLUX.2-klein-9B~\cite{flux-2-2025} with prompts from DiffusionDB~\cite{wangDiffusionDBLargescalePrompt2022}. Pure text data extracted from LLaVA-UHD-v3~\cite{sun2025llava}. The ratio of understanding, generation, and text data is $3:6:1$.

\noindent\textbf{Stage III: Refined Pre-Training.} We focus on visual reasoning and semantic alignment using 33M samples in this stage, maintaining a $3:6:1$ ratio across understanding, generation, and text data. We combine LLaVA-UHD-v3 instruction data~\cite{sun2025llava} for understanding and synthetic data generated via FLUX.2-klein-9B~\cite{flux-2-2025}, utilizing prompts from DiffusionDB~\cite{wangDiffusionDBLargescalePrompt2022} and LLaVA-OneVision-1.5~\cite{an2025llavaonevision15fullyopenframework}. To improve compositional reasoning (\eg, counting, color, and space), we also produced 466K instructions based on Objects365~\cite{shao2019objects365} to synthesize images. Pure text data is extracted from Nemotron-Cascade~\cite{wang2025nemotron}.

\noindent\textbf{Stage IV: Supervised Fine-Tuning.} 
We fine-tune the model on 3.8M curated samples, incorporating a high-quality subset of the Stage III data with Echo-4o-Image~\cite{ye2025echo}, MoviePosters~\cite{skvarre2023movie-posters-100k}, and ShareGPT-4o-Image~\cite{chen2025sharegpt}. During training, we maintain a $1:1$ batch ratio between understanding and generation tasks.

\section{Experiments}
\label{sec:experiments}
% We conduct extensive experiments to evaluate the effectiveness of \OurModelName{} from multiple perspectives, covering both visual understanding and generation tasks. We first introduce the evaluation setup in~\cref{sec:setup}, then present the main results on a wide range of multimodal benchmarks in~\cref{sec:main_results}, demonstrating the strong and consistent performance of \OurModelName{}. Furthermore, we perform detailed ablation studies to analyze the contributions of key components in \OurModelName{} in~\cref{sec:ablation}. To provide deeper insights, we include qualitative visualizations in~\cref{sec:visualization} that illustrate the model behavior and generation characteristics. Finally, we discuss additional findings and observations to further understand the strengths and limitations of our approach.
We evaluate \OurModelName{} on diverse multimodal benchmarks. We first describe the setup in~\cref{sec:setup} and report main results in~\cref{sec:main_results}. Subsequent analyses include visualizations in~\cref{sec:visualization}, ablation studies in~\cref{sec:ablation}, and an in-depth discussion of the model's characteristics and limitations.

\subsection{Evaluation Setup}
\label{sec:setup}
\noindent\textbf{Multimodal Understanding.}
We evaluate \OurModelName{} on diverse and widely recognized multimodal understanding benchmarks. (1) General Benchmarks: SEEDBench~\cite{li2023seed}, MMStar~\cite{chen2024we}, MMBench~\cite{liu2024mmbench}. (2) OCR Benchmarks: ChartQA~\cite{masry2022chartqa}, OCRBench~\cite{liu2024ocrbench}. (3) Visual Spatial Benchmarks: RealWorldQA~\cite{Grok15v}, POPE~\cite{li2023evaluating}. (4) Knowledge-focused Benchmarks: AI2D~\cite{kembhavi2016diagram}, MathVista~\cite{lu2023mathvista}, MMMU~\cite{yue2024mmmu}.

\noindent\textbf{Visual Generation.}
We evaluate visual generation performance on GenEval~\cite{ghosh2023geneval} and DPG-Bench~\cite{hu2024ella}. GenEval is an object-focused evaluation framework designed to rigorously assess the compositional alignment and fine-grained controllable generation capabilities of text-to-image models. DPG-Bench is a comprehensive benchmark comprising over a thousand dense prompts designed to evaluate the semantic alignment and prompt-following capabilities of text-to-image models in complex, multi-entity scenarios.
\subsection{Main Results}
\label{sec:main_results}

\begin{table}[t!]
\caption{Evaluation on multimodal understanding benchmarks. \#Params.: LLM backbone parameters.}
\label{tab:comprehension_result}
\centering
\resizebox{\textwidth}{!}{
\setlength{\tabcolsep}{2pt} % 默认6pt，可根据需要调成3~4pt，投稿前确认能不能,下同\hspace{6pt}
\begin{tabular}{@{\hspace{2pt}}lccccccccccc@{\hspace{2pt}}}
\toprule
\multirow{2}{*}{\textbf{Model}}&\multirow{2}{*}{\textbf{\#Params.}}&\multicolumn{3}{c}{\textbf{General}}&\multicolumn{2}{c}{\textbf{OCR}}&\multicolumn{2}{c}{\textbf{Visual Spatial}}&\multicolumn{3}{c}{\textbf{Knowledge}}\\
\cmidrule(lr){3-5}\cmidrule(lr){6-7}\cmidrule(lr){8-9}\cmidrule(lr){10-12}
&&\textbf{SEEDBench}&\textbf{MMStar}&\textbf{MMBench}&\textbf{ChartQA}&\textbf{OCRBench}&\textbf{RealWorldQA}&\textbf{POPE}&\textbf{AI2D}&\textbf{MathVista}&\textbf{MMMU}\\
\midrule
\multicolumn{12}{l}{\textbf{\textit{Understanding Only}}}\\ 
MobileVLM-V2~\cite{chu2024mobilevlm}&1.4B&-&-&57.7&-&-&-&84.3&-&-&-\\
Qwen2-VL~\cite{wang2024qwen2}&2B&-&48.0&72.2&73.5&80.9&62.9&-&74.7&43.0&41.1\\
DeepSeek-VL~\cite{lu2024deepseek}&7B&70.4&-&73.2&-&45.6&-&88.1&-&36.1&36.6\\
mPLUG-Owl2~\cite{ye2024mplug}&7B&57.8&-&64.5&22.8&25.5&50.3&86.2&55.7&-&32.7\\
\midrule
\multicolumn{12}{l}{\textbf{\textit{Understanding} \& \textit{Generation}}}\\ 
Emu3~\cite{wang2024emu3}&8B&68.2&-&58.5&\underline{68.6}&\textbf{68.7}&\underline{57.4}&85.2&\underline{70.0}&-&31.6\\
Show-o~\cite{xie2024show}&1.3B&51.5&-&-&-&-&-&80.0&-&-&26.7\\
Show-o2~\cite{xie2025show}&1.5B&65.6&\underline{43.4}&67.4&40.0&24.5&56.5&-&69.0&-&\underline{37.1}\\
JanusFlow~\cite{ma2025janusflow}&1.3B&\underline{70.5}&40.6&\underline{74.9}&64.6&53.2&41.2&\underline{88.0}&54.2&-&29.3\\
Janus-Pro~\cite{chen2025janus}&1.5B&68.3&43.1&\textbf{75.5}&23.4&48.7&52.6&86.2&64.5&-&36.3\\
Harmon~\cite{wu2025harmonizing}&1.5B&67.1&35.3&65.5&29.8&11.2&49.8&87.6&57.0&-&\textbf{38.9}\\
Tar~\cite{han2025vision}&1.5B&70.4&-&65.6&-&-&-&\textbf{88.4}&-&-&36.0\\
\cellcolor{lightpurple}\OurModelName{}&\cellcolor{lightpurple}1.5B&\cellcolor{lightpurple}\textbf{71.7}&\cellcolor{lightpurple}\textbf{50.9}&\cellcolor{lightpurple}70.4&\cellcolor{lightpurple}\textbf{75.7}&\cellcolor{lightpurple}\underline{58.4}&\cellcolor{lightpurple}\textbf{60.9}&\cellcolor{lightpurple}87.9&\cellcolor{lightpurple}\textbf{74.4}&\cellcolor{lightpurple}\textbf{50.5}&\cellcolor{lightpurple}36.0\\
\bottomrule
\end{tabular}
}
\end{table}
\noindent\textbf{Image Understanding.}
As shown in~\cref{tab:comprehension_result}, \OurModelName{} achieves competitive performance on nearly all benchmarks, demonstrating its strong and reliable understanding ability.

% \begin{table}[tb]
%   \caption{Performances of T2I generation ability on GenEval. "Und." and "Gen." denote "Understanding" and "Generation" respectively. "Obj." and "Attri." denote "Object" and "Attribute" respectively. "\#Params." indicates the number of parameters of LLM backbone.
%   }
%   \label{tab:main_results}
%   \centering
%   \resizebox{\textwidth}{!}{
%   \setlength{\tabcolsep}{4pt} % 默认6pt，可根据需要调成3~4pt，投稿前确认能不能用，下同\hspace{6pt}
%   \begin{tabular}{@{\hspace{6pt}}llccc@{\hspace{6pt}}}
%     \toprule
%     \multirow{2}{*}{\textbf{Model}} & \multirow{2}{*}{\textbf{LLM Backbone}} & \multicolumn{3}{c}{\textbf{Generation}} \\
%     \cmidrule(lr){3-5} & & \textbf{GenEval} & \textbf{DPG-Bench} & \textbf{gFID} \\
%     \midrule
%     Janus-Pro-1B & - & 0.73 & - & - \\
%     \OurModelName{} & - & - & - & - \\
%   \bottomrule
%   \end{tabular}
%   }
% \end{table}
%Performances of T2I generation ability on GenEval. ``Obj.'' and ``Attr.'' denote ``Object'' and ``Attribute'' respectively. ``\#Params.'' indicates the number of parameters of LLM backbone. ``\#Data'' indicates the total number of training samples, including visual understanding data, image generation data, and pure text data.
\begin{table}[th]
\caption{Performances on GenEval. Obj.: Object; Attr.: Attribute; \#Params.: LLM backbone parameters; \#Data: total training samples (visual, image, and text).}
\label{tab:geneval_result}
\centering
\resizebox{\textwidth}{!}{
\setlength{\tabcolsep}{4pt} % 默认6pt，可根据需要调成3~4pt，投稿前确认能不能用，下同\hspace{6pt}
\begin{tabular}{@{\hspace{6pt}}lcccccccc|c@{\hspace{6pt}}}
\toprule
\textbf{Model}&\textbf{\#Params.}&\textbf{\#Data}&\textbf{Single Obj.}&\textbf{Two Obj.}&\textbf{Counting}&\textbf{Colors}&\textbf{Position}&\textbf{Color Attr.}&\textbf{Overall}\\
\midrule
\multicolumn{8}{l}{\textbf{\textit{Generation Only}}}&\\ 
LlamaGen~\cite{sun2024autoregressive}&0.8B&62M&0.71&0.34&0.21&0.58&0.07&0.04&0.32\\
SDXL~\cite{podell2023sdxl}&2.6B&-&0.98&0.74&0.39&0.85&0.15&0.23&0.55\\
DALL-E 3~\cite{betker2023improving}&-&-&0.96&0.87&0.47&0.83&0.43&0.45&0.67\\
SD3-Medium~\cite{esser2024scaling}&2B&-&0.99&0.94&0.72&0.89&0.33&0.60&0.74\\%似乎未知
\midrule
% \noalign{\vskip -0.07cm} % Small vertical space
\multicolumn{8}{l}{\textbf{\textit{Understanding} \& \textit{Generation}}}&\\ 
% \noalign{\vskip -0.01cm} % Small vertical space
Chameleon~\cite{Chameleon_Team_Chameleon_Mixed-Modal_Early-Fusion_2024}&7B&5.2B&-&-&-&-&-&-&0.39\\
Show-o~\cite{xie2024show}&1.3B&3.2B&0.95&0.52&0.49&0.82&0.11&0.28&0.53\\
TokenFlow-XL~\cite{qu2025tokenflow}&14B&5.76B&0.95&0.60&0.41&0.81&0.16&0.24&0.55\\
Janus~\cite{wu2024janus}&1.3B&-&0.97&0.68&0.30&0.84&0.46&0.42&0.61\\
JanusFlow~\cite{ma2025janusflow}&1.3B&-&-&-&-&-&-&-&0.63\\
Transfusion~\cite{zhou2024transfusion}&7.3B&3.5B&-&-&-&-&-&-&0.63\\
D-DiT~\cite{li2025dual}&2B&40M&0.97&0.80&0.54&0.76&0.32&0.50&0.65\\
Emu3~\cite{wang2024emu3}&8B&-&0.99&0.81&0.42&0.80&0.49&0.45&0.66\\%似乎未知
Janus-Pro~\cite{chen2025janus}&1.5B&162M&0.98&0.82&0.51&0.89&0.65&0.56&0.73\\
Show-o2~\cite{xie2025show}&1.5B&177M&0.99&0.86&0.55&0.86&0.46&0.63&0.73\\
Harmon~\cite{wu2025harmonizing}&1.5B&113M&0.99&0.86&0.66&0.85&0.74&0.48&\underline{0.76}\\
Tar~\cite{han2025vision}&1.5B&403M&0.99&0.91&0.76&0.81&0.57&0.51&\underline{0.76}\\
\cellcolor{lightpurple}\OurModelName{}&\cellcolor{lightpurple}1.5B&\cellcolor{lightpurple}83M &\cellcolor{lightpurple}0.98&\cellcolor{lightpurple}0.92&\cellcolor{lightpurple}0.65&\cellcolor{lightpurple}0.86&\cellcolor{lightpurple}0.63&\cellcolor{lightpurple}0.65&\cellcolor{lightpurple}\textbf{0.78}\\
\bottomrule
\end{tabular}
}
\end{table}
% \begin{table}[tb]
%   \caption{Performances of T2I generation ability on GenEval. "Und." and "Gen." denote "Understanding" and "Generation" respectively. "Obj." and "Attri." denote "Object" and "Attribute" respectively. "\#Params." indicates the number of parameters of LLM backbone.
%   }
%   \label{tab:main_results}
%   \centering
%   \resizebox{\textwidth}{!}{
%   \setlength{\tabcolsep}{4pt} % 默认6pt，可根据需要调成3~4pt，投稿前确认能不能用，下同\hspace{6pt}
%   \begin{tabular}{@{\hspace{6pt}}llccc@{\hspace{6pt}}}
%     \toprule
%     \multirow{2}{*}{\textbf{Model}} & \multirow{2}{*}{\textbf{LLM Backbone}} & \multicolumn{3}{c}{\textbf{Generation}} \\
%     \cmidrule(lr){3-5} & & \textbf{GenEval} & \textbf{DPG-Bench} & \textbf{gFID} \\
%     \midrule
%     Janus-Pro-1B & - & 0.73 & - & - \\
%     \OurModelName{} & - & - & - & - \\
%   \bottomrule
%   \end{tabular}
%   }
% \end{table}
% Performances of dense prompt following ability on DPG-Bench. ``\#Params.'' indicates the number of parameters of LLM backbone. ``\#Data'' indicates the total number of training samples, including visual understanding data, image generation data, and pure text data.
\begin{table}[th]
\caption{Performances on DPG-Bench. \#Params.: LLM backbone parameters; \#Data: total training samples (visual, image, and text).}
\label{tab:dpg_result}
\centering
\resizebox{\textwidth}{!}{
\setlength{\tabcolsep}{6pt} % 默认6pt，可根据需要调成3~4pt，投稿前确认能不能用，下同\hspace{6pt}
\begin{tabular}{@{\hspace{6pt}}lccccccc|c@{\hspace{6pt}}}
\toprule
\textbf{Model}&\textbf{\#Params.}&\textbf{\#Data}&\textbf{Global}&\textbf{Entity}&\textbf{Attribute}&\textbf{Relation}&\textbf{Other}&\textbf{Overall}\\
\midrule
\multicolumn{7}{l}{\textbf{\textit{Generation Only}}}&\\ 
SDXL~\cite{podell2023sdxl}&2.6B&-&83.27&82.43&80.91&86.76&80.41&74.65\\
DALL-E 3~\cite{betker2023improving}&-&-&90.97&89.61&88.39&90.58&89.83&83.50\\
SD3-Medium~\cite{esser2024scaling}&2B&-&87.90&91.01&88.83&80.70&88.68&84.08\\%似乎未知
\midrule
% \noalign{\vskip -0.07cm} % Small vertical space
\multicolumn{7}{l}{\textbf{\textit{Understanding} \& \textit{Generation}}}&\\ 
% \noalign{\vskip -0.01cm} % Small vertical space
JanusFlow~\cite{ma2025janusflow}&1.3B&-&87.03&87.31&87.39&89.79&88.10&80.09\\
Emu3~\cite{wang2024emu3}&8B&-&85.21&86.68&86.84&90.22&83.15&80.60\\%似乎未知
Janus-Pro~\cite{chen2025janus}&1.5B&162M&87.58&88.63&88.17&88.98&88.30&82.63\\
Show-o2~\cite{xie2025show}&1.5B&177M&87.53&90.38&91.34&90.30&91.21&\textbf{85.02}\\
Tar~\cite{han2025vision}&1.5B&403M&83.59&89.35&86.91&93.50&80.80&82.96\\
\cellcolor{lightpurple}\OurModelName{}&\cellcolor{lightpurple}1.5B&\cellcolor{lightpurple}83M&\cellcolor{lightpurple}90.84&\cellcolor{lightpurple}90.24&\cellcolor{lightpurple}89.42&\cellcolor{lightpurple}89.71&\cellcolor{lightpurple}87.37&\cellcolor{lightpurple}\underline{83.48}\\
\bottomrule
\end{tabular}
}
\end{table}
\begin{figure}[tb]
\centering
\includegraphics[width=\linewidth]{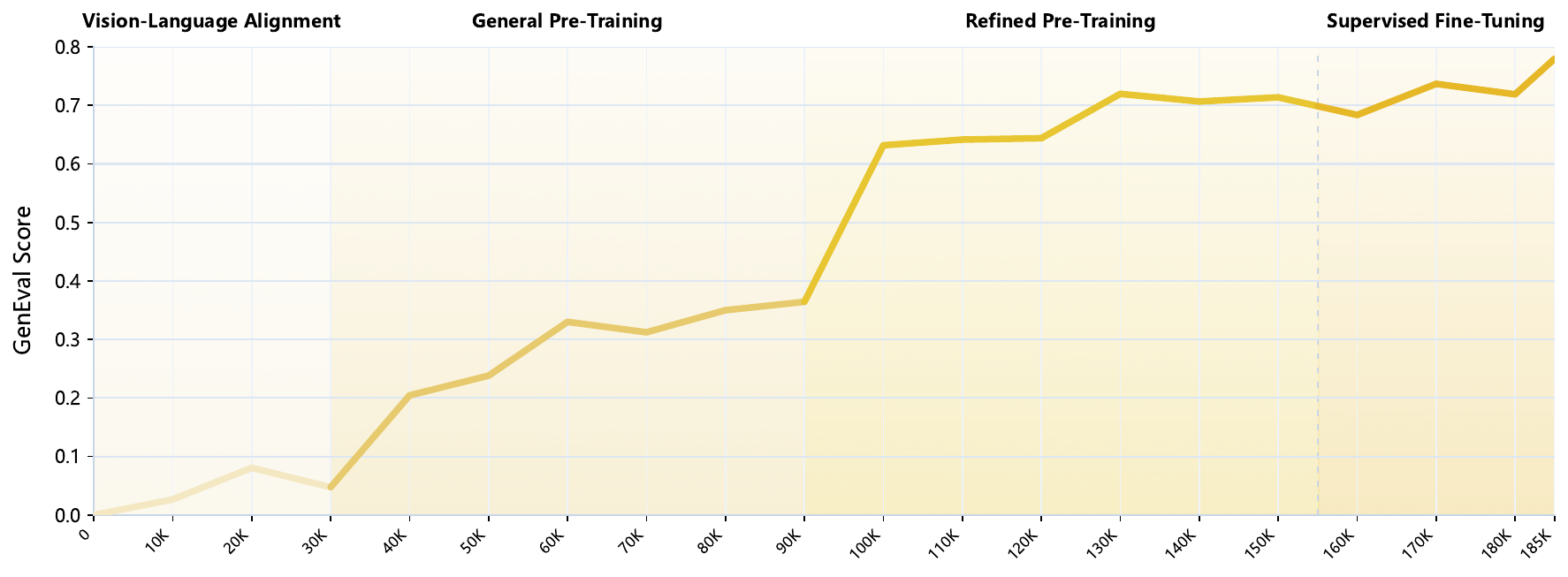}
\caption{Overall training pipeline and the progression of the GenEval score. The curve above illustrates the GenEval score as a function of cumulative training steps.}
\label{fig:geneval_step}
\end{figure}

\noindent\textbf{Image Generation.} 
% We evaluate the image generation capability of \OurModelName{} on two widely used benchmarks, GenEval and DPG-Bench. 
The results are summarized in~\cref{tab:geneval_result} and~\cref{tab:dpg_result}. Across all benchmarks, \OurModelName{} consistently achieves competitive or superior performance compared with existing approaches under comparable parameter scales, including models such as Janus-Pro. Notably, \OurModelName{} attains these strong results using only 83M training samples in total, demonstrating that high-quality generation does not rely solely on large-scale data. This highlights the effectiveness of our unified architecture, whose shared representation design enables efficient knowledge transfer between understanding and generation, resulting in robust image synthesis performance with high data efficiency.

\noindent\textbf{Progressive Improvement of Generation Capability.}
To illustrate how the generation capability of \OurModelName{} evolves throughout training, we present the progression of GenEval scores across different stages in~\cref{fig:geneval_step}. As shown in the figure, the model exhibits steady improvement as training proceeds, with clear performance gains at each stage. During the Vision-Language Alignment and General Pre-Training stages, most of the generation training data consists of real-world natural images paired with captions. Such data are complex and ambiguous, as real-world scenes often contain multiple objects, intricate interactions, and incompletely described visual details, making it difficult for the model to learn precise text-to-image correspondences. As a result, although the model gradually acquires fundamental visual semantic understanding during these stages, the generation performance improves moderately and remains sub-optimal until a significant surge in the Refined Pre-Training stage, where the generation training data are primarily synthetic and instruction-oriented. Compared with real-world data, synthetic data provides clearer object compositions, more explicit attribute bindings, and more direct text-image correspondence, making the learning objective better defined and easier to optimize. This significantly enhances generation fidelity and alignment, leading to rapid improvement in generation ability. Finally, during Supervised Fine-Tuning, we adopt a smaller learning rate together with a cosine decay schedule to stabilize optimization, reduce overfitting, and encourage smoother convergence. This stage further refines output quality and alignment consistency, yielding stable and incremental performance gains. Overall, these results demonstrate that the generation capability of \OurModelName{} improves in a progressive manner, validating the effectiveness of our staged training pipeline.
\subsection{Analysis of High-Frequency Injection}
\label{sec:visualization}
\begin{figure}[th]
\centering
\includegraphics[width=\linewidth]{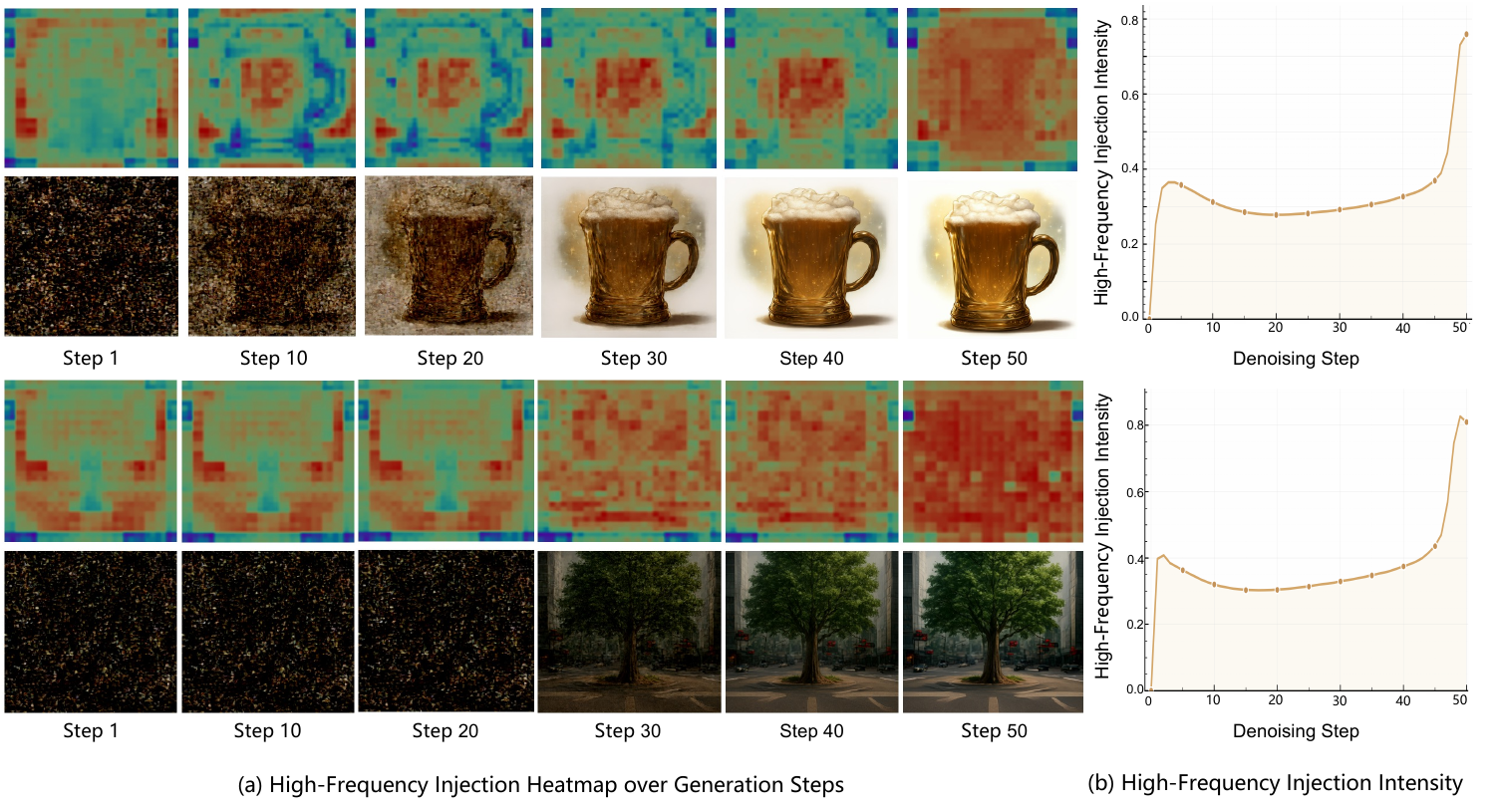}
\caption{(a)~Heatmap of high frequency injection across different generation steps. (b)~High frequency injection intensity at each step. The reported values are aggregated over multiple runs, with per-run normalization applied prior to averaging across samples at the same step, ensuring a faithful representation of the overall trend.}
\label{fig:denoise}
\end{figure}

\noindent To investigate how high-frequency information contributes throughout the generation process, we visualize the high-frequency injection patterns across different denoising steps, as shown in~\cref{fig:denoise}~(a). The heatmaps reveal a clear temporal structure. At the early stage of generation, high-frequency components are sparsely activated and mainly concentrate around the formation of the primary object contours. As generation progresses to the intermediate stage, the magnitude of HFI slightly decreases, and the model relies primarily on semantic and low-frequency signals to complete structural details and object-level compositions. In the final stage, when the overall image structure has largely stabilized, the activation of high-frequency components increases significantly, contributing to the refinement of local textures and fine-grained visual details. This stage-wise evolution suggests that high-frequency signals are not uniformly involved, but instead dynamically modulated according to the generation stage.

\cref{fig:denoise}~(b) further quantifies the degree of high-frequency participation across the entire denoising trajectory. At the beginning of generation, the model focuses on constructing coarse layouts and global structures, during which HFI starts from a relatively low level. In the middle phase, semantic and low-frequency information guide the generation of object-level structures and attributes, resulting in a relatively moderate level of high-frequency participation. As the process approaches the final stages, the intensity and proportion of HFI increase sharply, indicating its growing role in enhancing local textures and visual fidelity. Such a progressive pattern reflects a hierarchical generation mechanism, where the model first sketches global layouts and semantic structures before gradually refining localized details and textures, resembling the coarse-to-fine process commonly observed in human drawing behavior.
\subsection{Ablation Studies}
\label{sec:ablation}
\noindent We conduct controlled ablation experiments to investigate HFI and the impact of the generation objective on multimodal understanding. After standard Vision–Language Alignment, models are fine-tuned using 858K understanding and 850K generation samples (randomly sampled from Refined Pre-Training) for preliminary investigation and rapid validation.

\noindent\textbf{Does Generation Affect Understanding?}
To further examine whether jointly training understanding and generation tasks under this architecture compromises multimodal understanding performance, we conduct a controlled experiment. After completing the alignment stage, we fine-tune a model using only multimodal understanding data as a baseline for comparison. The results, summarized in~\cref{tab:ablation}, show that joint training with both understanding and generation objectives not only equips the model with image generation capability, but also achieves comparable or even slightly superior understanding performance compared to fine-tuning on understanding data alone. These findings demonstrate the effectiveness of the unified vision tokenizer design in supporting multimodal tasks within a single framework.

\begin{table}[t]
\caption{Ablation results. The first row corresponds to a model fine-tuned using understanding data only. The second and third rows report jointly trained models optimized for both understanding and generation, without and with HFI, respectively.}
\label{tab:ablation}
\centering
\resizebox{\textwidth}{!}{
\setlength{\tabcolsep}{6pt}
\begin{tabular}{@{\hspace{2pt}}lcc|ccccc|cc@{\hspace{2pt}}}
\toprule
\textbf{Model}&\textbf{HFI}&\textbf{Fine-tuning Data}&\textbf{SEEDBench}&\textbf{MMBench}&\textbf{ChartQA}&\textbf{POPE}&\textbf{AI2D}&\textbf{GenEval}&\textbf{DPG-Bench}\\
\midrule
\OurModelName{}&\cmark&Understanding&70.8&65.2&58.5&87.0&67.7&-&-\\
\OurModelName{}&\xmark&Generation \& Understanding&70.0&66.3&58.8&86.2&67.3&0.17&39.11\\
\OurModelName{}&\cmark&Generation \& Understanding&69.8&67.1&59.9&87.5&68.1&0.30&51.63\\
\bottomrule
\end{tabular}
}
\end{table}

\noindent\textbf{Is High-Frequency Injection Necessary?}
As shown in~\cref{tab:ablation}, we train two models under identical training settings: \OurModelName{} and a variant without HFI. The results indicate that introducing high-frequency patch details has minimal impact on multimodal understanding performance, while leading to a substantial improvement in generation quality. Although the model without HFI is able to produce semantically consistent images, the generated results lack fine-grained visual details. In contrast, incorporating high-frequency patch details significantly enhances texture fidelity and structural sharpness. These findings suggest that HFI plays a crucial role in improving visual detail generation, making it an essential component for unified multimodal modeling.
\subsection{Discussion}
\label{sec:discussion}
%   Vision encoder 可继承pretrained model without tokenizer pretraining 【继承原生VLM】
%   Vae-decoder ocr-related 性能 @yichen 填表
\OurModelName{} directly leverages the pre-trained weights of a native ViT model (\eg, SigLIP2), allowing the model to fully benefit from the rich representations learned during pre-training. In this way, we avoid the computational overhead of training a unified vision encoder, which has long been considered a key step in prior work for achieving unified vision-language understanding and generation. Moreover, our architecture does not freeze any core parameters, enabling all modules to be jointly fine-tuned. This ensures maximal synergy among components and allows the model to fully exploit the strengths of each module.

\section{Related Work}
\label{sec:related_work}

\subsection{Image Tokenizers.}
% 这段可以参考tuna 论文里面的设计原则那段写。就是总分结构，
% 从离散和连续角度写一段
% 从语义和细节写一段。
% 统一的比如rae、svg、tar的tokenzier
Unified Multimodal Models (UMMs) necessitate a versatile visual representation capable of bridging the gap between discriminative understanding and generative synthesis. The design of image tokenizers is central to this objective, revolving around the trade-offs between representation formats, feature granularities, and architectural integration.

\noindent\textbf{Discrete \vs Continuous Representations.} 
The choice of tokenization dictates the modeling paradigm. Discrete tokenizers like Chameleon~\cite{Chameleon_Team_Chameleon_Mixed-Modal_Early-Fusion_2024} map images to a finite codebook to leverage autoregressive objectives, yet often suffer from information bottlenecks and reduced fidelity. In contrast, continuous representations in KL-regularized latent spaces preserve richer details, becoming the preference for high-fidelity generation. As highlighted in TUNA~\cite{liu2025tuna}, continuous features, when aligned with pretrained encoders, exhibit superior efficacy for both generative and semantic understanding tasks.

\noindent\textbf{Semantic Features \vs High-Frequency Textures.}
A fundamental disparity exists between low-frequency semantics for reasoning and high-frequency textures for reconstruction. Semantic encoders like SigLIP~\cite{tschannen2025siglip} often overlook fine-grained textures, leading to blurred synthesis, while generative VAEs lack global context. To reconcile this, Show-o~\cite{xie2024show} explores late-fusion strategies, while TUNA~\cite{liu2025tuna} proposes a cascaded architecture that extracts semantics directly from VAE latents to achieve a balanced and unified representation space.

\noindent\textbf{Unified visual tokenizer.}
Recent efforts focus on native unification. Discrete approaches like UniTok~\cite{ma2025unitok} and TokLIP~\cite{lin2025toklip} attempt to learn shared codebooks for dual tasks but remain constrained by discretization limits. Conversely, continuous unified frameworks integrate representation encoders with generative spaces. For instance, RAE~\cite{zheng2025diffusion,tong2026scaling} leverages frozen semantic features for reconstruction, while TAR~\cite{han2025tar} and SVG~\cite{shi2025latent} explore joint modeling. However, achieving a seamless balance between high-level reasoning and pixel-perfect generation remains a core challenge in UMM design.

\subsection{Unified Multimodal Models.}
% 这段可以参考tuna 论文里面的设计原则那段写
% 总分结构，先说有什么样的umms，分为纯ar的、分为纯diffusion的、分为ar+diffusion的
% （1）ar 
% （2）diff 
% （3）ar+diffusion
Unified Multimodal Models (UMMs) represent a paradigm shift toward native architectural integration, aiming to bridge the structural divide between multimodal perception and generative synthesis within a shared transformer backbone. Current research can be broadly categorized into three primary paradigms based on their modeling objectives: pure autoregressive (AR), pure diffusion, and hybrid frameworks. Pure AR models, exemplified by Chameleon~\cite{Chameleon_Team_Chameleon_Mixed-Modal_Early-Fusion_2024}, Emu3~\cite{wang2024emu3}, and Janus-Pro~\cite{chen2025janus}, unify modalities by quantizing visual data into discrete token sequences and optimizing via the next-token prediction objective. This approach allows them to inherit the proven scaling laws and reasoning capabilities of large language models. Conversely, pure diffusion-based UMMs such as MMaDA~\cite{yang2025mmada} and UniDisc~\cite{swerdlow2025unified} adopt unified masked token prediction or stochastic denoising. These models benefit from parallel decoding efficiency, which achieves significantly higher inference speeds than sequential AR, and support bidirectional reasoning for tasks like joint image-text inpainting. Finally, hybrid architectures strategically fuse these mechanisms to maximize cross-modal synergy. Models like Show-o~\cite{xie2024show}, Transfusion~\cite{zhou2024transfusion}, and SEED-X~\cite{ge2024seed} typically maintain autoregressive modeling for sequential language while employing diffusion or flow-matching processes for continuous visual representations, thereby achieving high-fidelity generation without compromising linguistic logic.

\subsection{Synergy in Multimodal Comprehension and Generation.}
% 这块来写理解生成协同效应
% 主要写理解生辅助生成【比如cot、scaling rae中的test time 增强、以及xiesaining的REPA，还有VAVAE】
% 写生成辅助理解：Ross【一篇利用重构任务提升理解的工作】，https://arxiv.org/pdf/2601.21406 这篇论文中有挺多的
The bidirectional synergy between multimodal comprehension and generation has become a pivotal research focus. Recent studies demonstrate that discriminative understanding significantly benefits generative performance. Specifically, REPA~\cite{yu2024representation} aligns diffusion transformer features with pretrained visual encoders to accelerate convergence, while VA-VAE~\cite{yao2025reconstruction} addresses the optimization dilemma between reconstruction and generation by aligning latent spaces with vision foundation models. Conversely, generative tasks increasingly serve to bolster visual comprehension. ROSS~\cite{wang2024reconstructive} introduces a reconstructive objective via latent denoising to enhance fine-grained perception and reduce hallucinations. Furthermore, UniMRG~\cite{su2026generation} incorporates auxiliary generation of intrinsic representations such as depth and segmentation to capture geometric and structural cues. These advancements suggest that internalizing generative tasks allows unified models to develop a more comprehensive understanding of spatial relations and structural layouts.
\section{Conclusion}
\label{sec:conclusion}
In this work, we present \OurModelName{}, a novel unified multimodal model that successfully harmonizes visual comprehension and high-fidelity generation within a single framework. The core of our approach lies in the decoupling of patch-level details from semantic representations, addressing the intrinsic optimization conflict that often plagues joint multimodal modeling. By utilizing a unified vision tokenizer to extract stable semantics and a cascaded flow matching head to inject semantically gated high-frequency residuals, \OurModelName{} ensures both robust understanding and precise image synthesis. Extensive evaluations across ten understanding benchmarks and multiple generation benchmarks demonstrate that \OurModelName{} achieves competitive performance. Notably, our model maintains high efficiency through a 4$\times$ token compression rate and shows the emergence of zero-shot image editing capabilities, despite being trained on a relatively modest dataset of 83M samples. These results validate that the hierarchical progression from global semantic layout to localized detail refinement—resembling the human drawing process—is an effective paradigm for unified modeling. While \OurModelName{} exhibits strong performance, future research could explore further scaling of both the LLM backbone and training data to unlock even more complex reasoning and creative generation abilities. Additionally, extending this decoupled representation framework to video understanding and generation presents a promising avenue for achieving more generalized multimodal intelligence.

\noindent\textbf{Limitation.} 
Despite its performance, our study has three primary constraints. First, the relatively small parameter scale of \OurModelName{} may limit its ability to capture intricate details. Second, as it is not initialized from large-scale pre-trained VLMs, its inherent visual understanding and generation capabilities require further enhancement. Finally, the current training pipeline relies on single-image datasets; future work will incorporate more diverse and complex multimodal data to improve generalization.

%Bibliography
\bibliographystyle{unsrt}  
\bibliography{references}  

\clearpage
\appendix
\section{Why Reconstruct Pixels Before Semantic Encoding?}
\begin{table}[h]
\caption{Effect of Pixel-Reconstruct on OCR-centric understanding and Others. We scale the alignment dataset from 558K to 4.6M samples (both fine-tuned on the same 858K data) to ensure sufficient training, and then introduce Pixel-Reconstruct under the same 4.6M alignment setting for comparison.}
\label{tab:ocr_suppl}
\centering
\resizebox{\textwidth}{!}{
\setlength{\tabcolsep}{6pt}
\begin{tabular}{@{\hspace{2pt}}lcc|cccccc@{\hspace{2pt}}}
\toprule
\textbf{Model}&\textbf{Pixel-Reconstruct}&\textbf{Training Data}&\textbf{ChartQA}&\textbf{OCRBench}&\textbf{TextVQA}&\textbf{MMBench}&\textbf{MMStar}&\textbf{SEEDBench}\\
\midrule
\OurModelName{}&\xmark&558K \& 858K&13.9&2.5&9.6&23.2&27.7&38.4\\
\OurModelName{}&\xmark&4.6M \& 858K&14.2&2.2&9.5&42.8&32.1&56.9\\
\OurModelName{} (Ours)&\cmark&558K \& 858K&42.1&31.5&43.4&66.09&40.2&69.2\\
\bottomrule
\end{tabular}
}
\end{table}

In our initial design, the latent representation produced by the VAE encoder was directly projected and fed into the SigLIP2-ViT for semantic encoding. However, preliminary experiments revealed that this design leads to severe degradation in OCR-centric understanding ability. Here we conducted a controlled study to analyze this issue.

All experiments in this section are conducted at a resolution of $256 \times 256$, focusing only on evaluating the model’s visual understanding capability. The first two rows in Table~\ref{tab:ocr_suppl} follow the design adopted in TUNA, where the latent representation from the VAE encoder is directly processed by a ViT backbone without using a VAE decoder. Under this setting, the model exhibits extremely poor performance on OCR-related benchmarks, indicating that fine-grained textual information is largely lost. To verify whether this issue stems from insufficient training data, we further scale the alignment dataset to 4.6M samples while keeping the same 858K fine-tuning set. As shown in the second row of Table~\ref{tab:ocr_suppl}, increasing the training data brings negligible improvement on OCR-related benchmarks, suggesting that the performance degradation is not caused by insufficient data but rather by the architectural design.

To address this issue, we introduce a VAE decoder that reconstructs the latent representation back into the pixel space before semantic encoding. The reconstructed image is then processed by the SigLIP2-ViT to extract semantic tokens. As shown in the third row of Table~\ref{tab:ocr_suppl}, introducing the reconstruction stage leads to substantial improvements across OCR-centric benchmarks. These results indicate that reconstructing pixels before semantic encoding is necessary for preserving fine-grained visual details that are crucial for text recognition.
\section{Emergent Abilities}
\begin{table}[ht]
\caption{Emergent capabilities of \OurModelName{} after Refined Pre-Training.}
\label{tab:tab_emerge}
\centering
\fontsize{10}{10}\selectfont
\resizebox{\linewidth}{!}{
\begin{tabular}{p{\linewidth}}
\toprule
\textbf{Change the color of the background to blue.} \\
\begin{minipage}{\textwidth}
\centering
\includegraphics[width=0.2\linewidth]{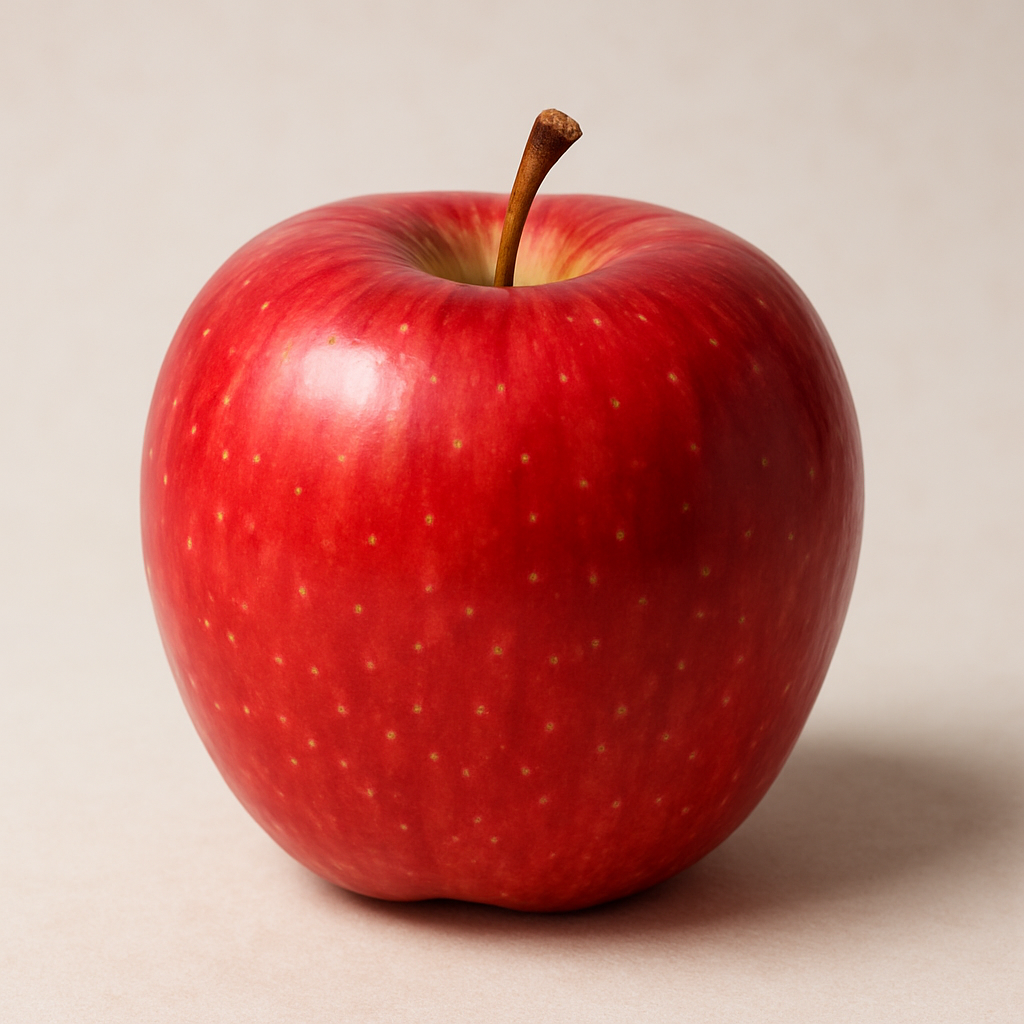} 
\hspace{0.02\linewidth}
\includegraphics[width=0.2\linewidth]{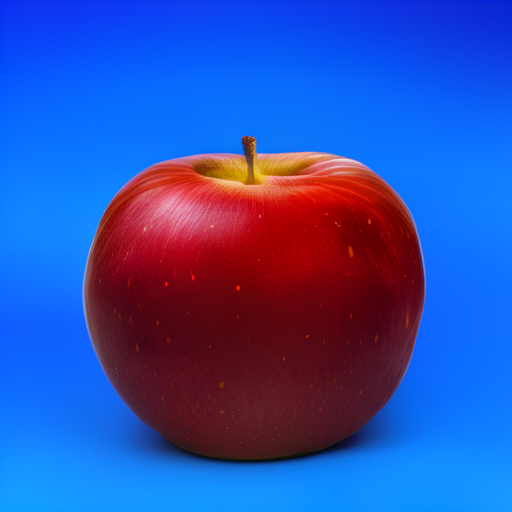} \\
\end{minipage}
\\
\midrule
\textbf{Turn this apple into a watermelon.} \\
\begin{minipage}{\textwidth}
\centering
\includegraphics[width=0.2\linewidth]{tab_suppl/figs/1.jpg} 
\hspace{0.02\linewidth}
\includegraphics[width=0.2\linewidth]{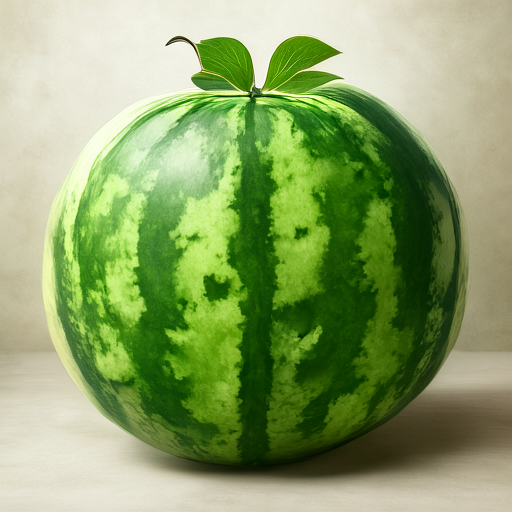} \\
\end{minipage}
\\
\midrule
\textbf{Place the red bell pepper from Figure 1 and the green broccoli from Figure 2 into one image.} \\
\begin{minipage}{\textwidth}
\centering
\includegraphics[width=0.2\linewidth]{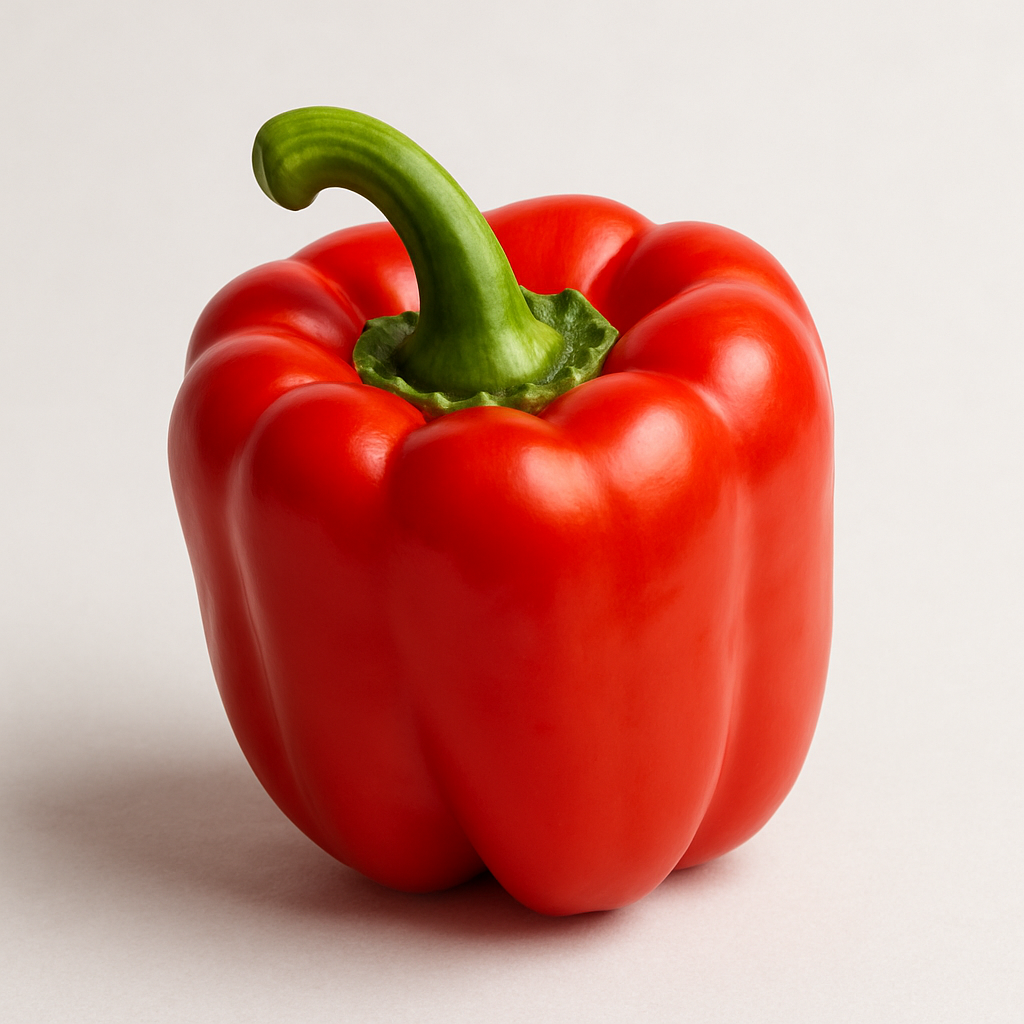} 
\hspace{0.02\linewidth}
\includegraphics[width=0.2\linewidth]{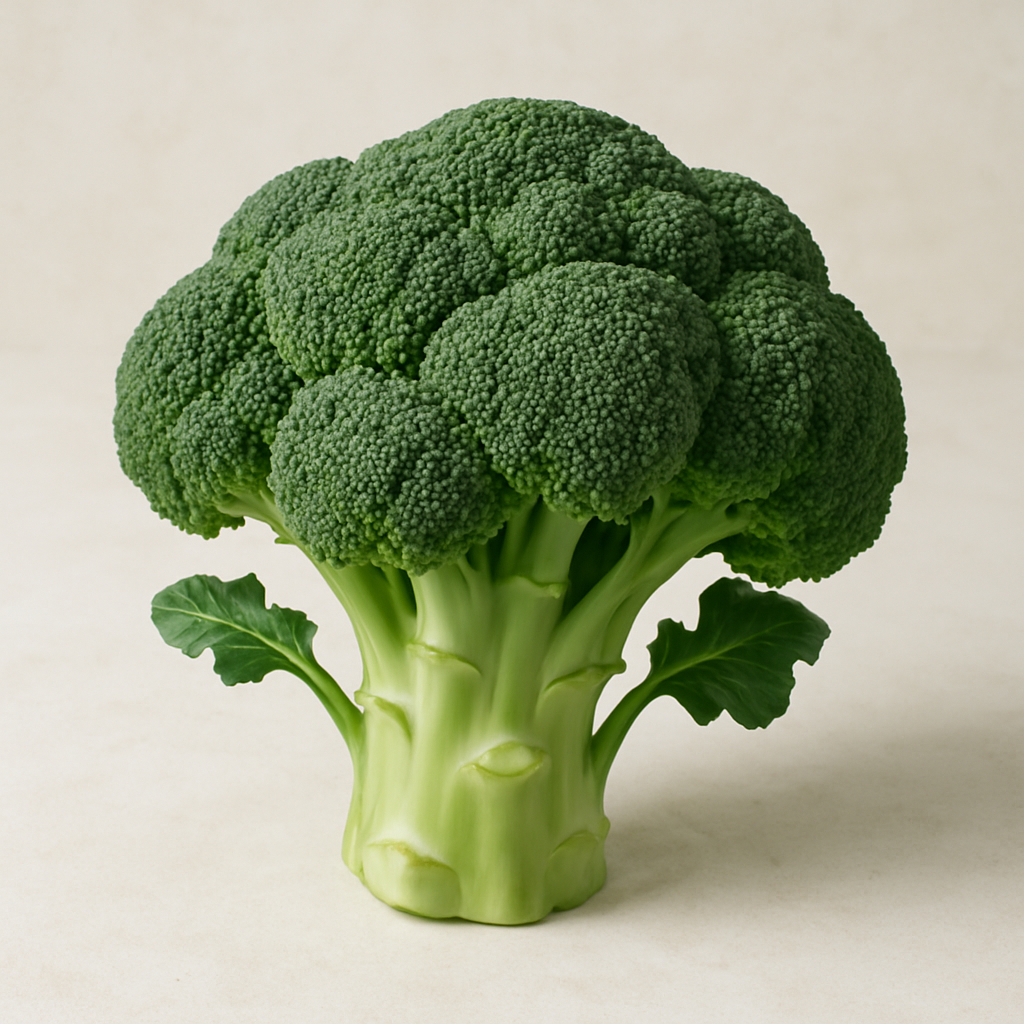} 
\hspace{0.02\linewidth}
\includegraphics[width=0.2\linewidth]{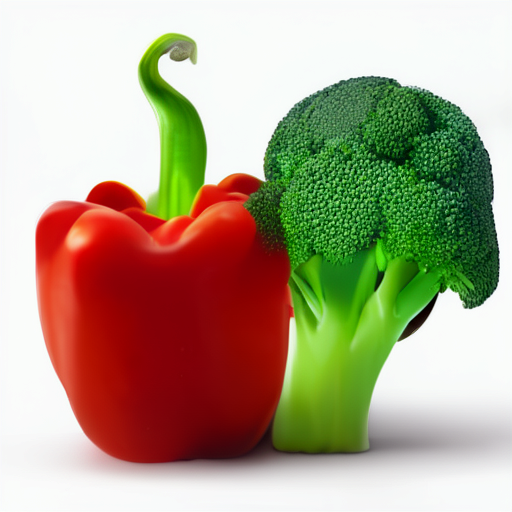} \\
\end{minipage}
\\
\bottomrule
\end{tabular}
}

\end{table}
To investigate the advantages of our Unified Visual Tokenizer in learning a shared feature space, we evaluate the checkpoint after Stage~3 training and observe several intriguing emergent capabilities. Notably, throughout all training stages up to Stage~3, the generation training only involves text-to-image tasks, without exposure to any multi-image or image editing data.

As show in~\cref{tab:tab_emerge}, despite not being trained on any image editing or multi-image data, \OurModelName{} still demonstrates remarkable potential on these tasks. This observation highlights the strong generalization ability of the proposed framework. In particular, the unified visual tokenizer enables different visual tasks to be represented within a shared feature space, which naturally facilitates capability transfer across tasks. These results further validate the effectiveness of learning a unified visual representation.
\section{Examples of Multimodal Understanding}
\begin{table}[b!]
\caption{Emergent capabilities of \OurModelName{} after Refined Pre-Training.}
\label{tab:tab_understanding}
\centering \fontsize{10}{10}\selectfont 
\resizebox{\linewidth}{!}{ 
\begin{tabular}{p{\linewidth}} 
\toprule 
\textbf{Discribe this image.}

\begin{wrapfigure}[14]{l}{0.3\linewidth}
\vspace{-12pt}
\centering
\includegraphics[width=\linewidth]{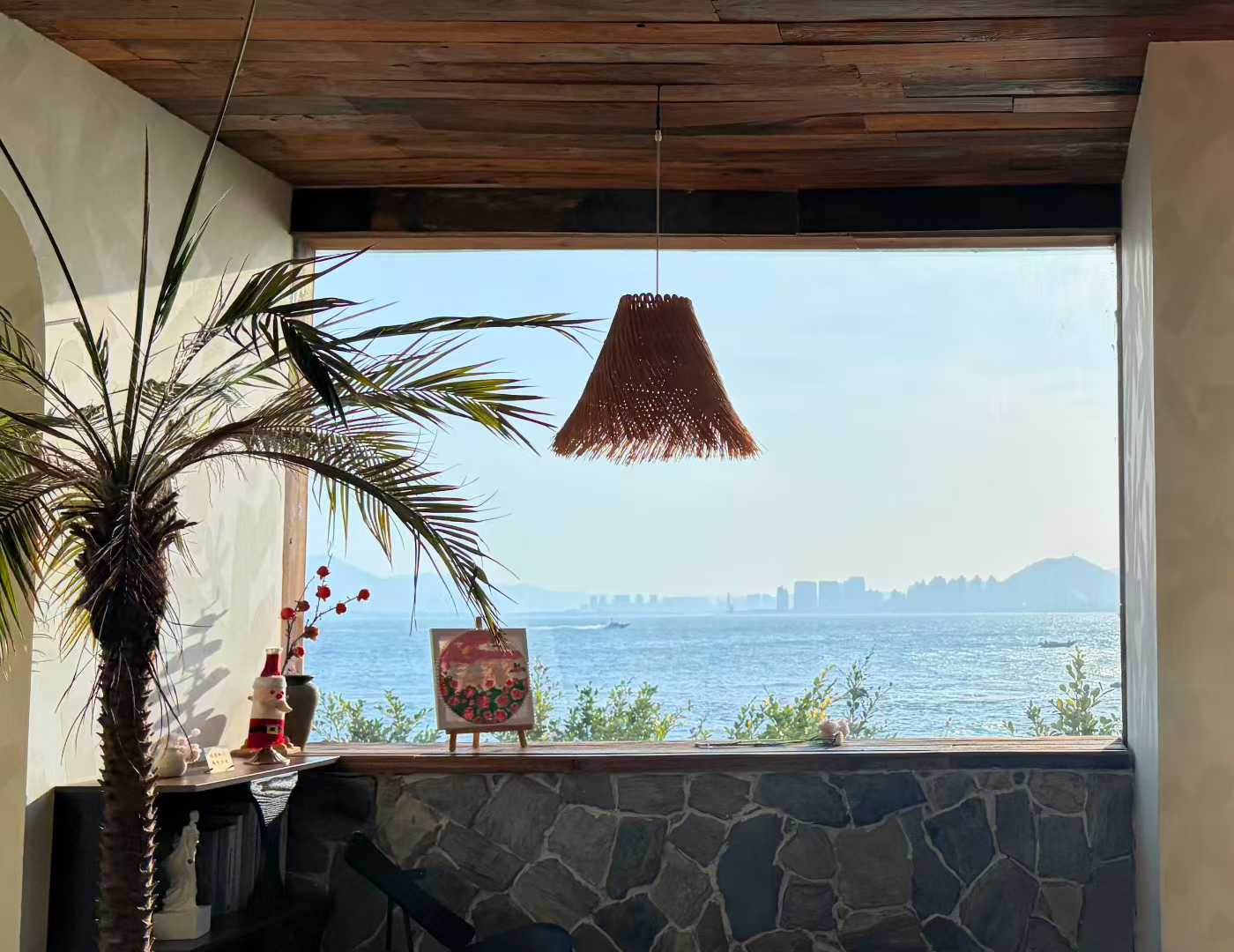}
\end{wrapfigure}
This image captures a cozy corner of a room with a stunning view of the ocean. The foreground features a stone wall with a wooden ceiling, creating a warm and rustic atmosphere. A black chair is tucked under a small table, which holds a vase with red flowers and a framed picture. large window dominates the scene, offering an impressive view of the sea and distant land. The ocean stretches out to the horizon, meeting a clear blue sky. In the distance, a city skyline and mountains can be seen, adding depth and interest to the landscape. Inside the room, a straw-like hanging lamp adds a touch of texture and soft lighting. A potted palm tree adds a tropical feel to the space. The overall scene is serene and inviting, with a perfect blend of natural and man-made elements. It's a peaceful coastal retreat that combines comfort with a breathtaking view. \\
\midrule
\textbf{Please answer the question requiring a floating-point number with one decimal place and provide the final value, \eg, 1.2, 1.3, 1.4, at the end. Question: What is the difference of largest and smallest bar?}

\begin{wrapfigure}[14]{l}{0.3\linewidth}
\vspace{-12pt}
\includegraphics[width=\linewidth]{tab_suppl/figs/case2.jpg}
\end{wrapfigure}
To find the difference between the largest and smallest bar, we need to identify the values of the largest and smallest bars from the given data. The largest bar represents Iceland, with a value of 100.0. The smallest bar represents Cambodia, with a value of 52.4. Now, we need to calculate the difference between these two values: 100.0 - 52.4 = 47.6 Therefore, the difference between the largest and smallest bar is 47.6. \\
\\
\\
\\
\\
\bottomrule
\end{tabular}
}

\end{table}
The main paper reports the quantitative performance of \OurModelName{} on multimodal understanding tasks. To further illustrate its capabilities, we present several qualitative examples in~\cref{tab:tab_understanding}. These cases demonstrate that \OurModelName{} is able to effectively comprehend and reason over multimodal inputs, exhibiting strong multimodal understanding ability. For multimodal understanding, input images are resized so that the longer side is 512 pixels while preserving the aspect ratio, and then padded to a resolution of $512 \times 512$ before being fed into the model.
\end{document}